\pdfoutput=1

\documentclass[11pt]{article}

\usepackage[preprint]{acl}

\usepackage{times}
\usepackage{latexsym}

\usepackage[T1]{fontenc}

\usepackage[utf8]{inputenc}

\usepackage{microtype}

\usepackage{inconsolata}

\usepackage{graphicx}

\usepackage{booktabs}
\usepackage{paralist}
\usepackage{multicol}
\usepackage[color=white,size=footnotesize]{todonotes}
\usepackage{array}
\usepackage[most]{tcolorbox}
\usepackage{amsmath}
\usepackage{amssymb}
\usepackage{multirow}
\usepackage{siunitx}
\usepackage{algorithm}
\usepackage{algpseudocode}
\usepackage{listings}
\usepackage{xcolor}

\usepackage{todonotes}

\lstdefinestyle{python}{
    language=Python,
    basicstyle=\ttfamily\small,
    keywordstyle=\color{blue}\bfseries,
    commentstyle=\color{gray}\itshape,
    stringstyle=\color{red},
    showstringspaces=false,
    breaklines=true,
    breakatwhitespace=true,
    postbreak=\mbox{\textcolor{gray}{$\hookrightarrow$}\space},
    frame=single,
    numbers=left,
    numberstyle=\tiny\color{gray},
    numbersep=5pt,
    tabsize=2,
    captionpos=b,
    escapeinside={(*@}{@*)},
    xleftmargin=1em,
    xrightmargin=1em,
    aboveskip=1em,
    belowskip=1em,
    keepspaces=true
}
\title{Grammar Search for Multi-Agent Systems}

\author{
  \textbf{Mayank Singh}$^{1}$\thanks{Corresponding author: \texttt{mayanks43@arizona.edu}},
  \textbf{Vikas Yadav}$^{2}$,
  \textbf{Shiva Krishna Reddy Malay}$^{2}$,
  \textbf{Shravan Nayak}$^{3}$,\\
  \textbf{Sai Rajeswar}$^{2}$,
  \textbf{Sathwik Tejaswi Madhusudhan}$^{2}$,
  \textbf{Eduardo Blanco}$^{1}$\\[4pt]
  $^{1}$University of Arizona \quad
  $^{2}$ServiceNow \quad
  $^{3}$Mila - Quebec AI Institute
}

\begin{document}
\maketitle

\begin{abstract}
Automatic search for Multi-Agent Systems has recently emerged
as a key focus in agentic AI research.
Several prior approaches have relied on LLM-based free-form search
over the code space.
In this work, we propose a more structured framework that explores
the same space through a fixed set of simple, composable components.
We show that, despite lacking the generative flexibility of LLMs
during the candidate generation stage, our method outperforms prior approaches
on four out of five benchmarks across two domains:
mathematics and question answering.
Furthermore, our method offers additional advantages,
including a more cost-efficient search process and the generation of modular,
interpretable multi-agent systems with simpler logic.
\end{abstract}

\section{Introduction}
\label{s:introduction}

Large language models (LLMs) have emerged as powerful general-purpose reasoning
systems capable of solving diverse tasks across language understanding,
reasoning, and decision-making \cite{comanici2025gemini, openai2025gpt5}.
Increasingly, LLMs are deployed as autonomous agents
for real-world applications such as coding \cite{yang2024swe},
tool use \cite{schick2023toolformer}, and computer interaction \cite{he-etal-2024-webvoyager}.
This shift has inspired a growing line of work that organizes multiple LLMs
into \textit{multi-agent systems} (MASes),
where specialized agents collaborate, exchange information,
and refine each other's outputs to tackle complex problems
\cite{du2023improving, madaan2023self, yao2023tree}.
This paradigm reframes LLMs from being standalone solvers to components
in a coordinated reasoning framework.

Early MASes were manually constructed, with researchers designing agent roles,
communication patterns, and reasoning strategies
by hand \cite{li2023camel, 10.1145/3586183.3606763}.
While effective at small scales, such manual design is labor-intensive
and does not generalize across tasks
\cite{DBLP:journals/fcsc/WangMFZYZCTCLZWW24}.
To overcome these limitations,
recent work has focused on automatic MAS search,
where the goal is to discover effective MASes
with minimal human input \cite{zhuge2024gptswarm, zhang2025multi, yue2025masrouter}.
These approaches define a search space of MASes
and use optimization or exploration algorithms to navigate it.

\begin{figure*}[t]
\centering
  \begin{tikzpicture}[
    box/.style={rectangle, draw, thick, text width=2.2cm, text centered, rounded corners=8pt, minimum height=1cm, font=\small},
    longbox/.style={rectangle, draw, thick, text width=3.2cm, text centered, rounded corners=8pt, minimum height=1cm, font=\small},
    example/.style={rectangle, draw, dashed, text width=3.2cm, text centered, rounded corners=5pt, fill=gray!5, font=\scriptsize},
    arrow/.style={->, thick, >=stealth}
]

\node[box] (grammar) at (0,0) {General-Purpose Grammar for MASes};
\node[box] (base) at (0,-2) {Components identified from Base MASes};
\node[longbox] (explore) at (4,0) {Grammar Search};
\node[longbox] (evaluate) at (8,0) {Evaluate MASes w/ Validation Set};
\node[longbox] (output) at (12,0) {Select Best MAS};

\draw[arrow] (grammar) -- (explore);
\draw[arrow] (base) -- (grammar);
\draw[arrow] (explore) -- (evaluate);
\draw[arrow] (evaluate) -- (output);

\node[example, text width=5.1cm] at (4,-1.8) {
    \textbf{Generate MASes:}\\
    MAS\textsubscript{1}: \texttt{StepByStepReasoner}\\
    MAS\textsubscript{2}: \texttt{RoleBasedReasoner $\Rightarrow$ MajorityVoter}\\
    MAS\textsubscript{3}: \texttt{StepByStepReasoner $\Rightarrow$ MajorityVoter}\\
    ...
};

\node[example, text width=2.3cm] at (8,-1.8) {
    \textbf{Results:}\\
    MAS\textsubscript{1}: 72\%\\
    MAS\textsubscript{2}: 78\%\\
    MAS\textsubscript{3}: 81\%\\
    ...
};

\node[example, text width=5.1cm] at (12,-1.8) {
    \textbf{Best MAS:}\\
    MAS\textsubscript{3}: \texttt{StepByStepReasoner $\Rightarrow$ MajorityVoter}\\[3pt]
    Accuracy: 81\%
};

\draw[dashed, gray] (explore) -- (4,-1.2);
\draw[dashed, gray] (evaluate) -- (8,-1.2);
\draw[dashed, gray] (output) -- (12,-1.2);

\end{tikzpicture}
\caption{
  Overview of our grammar-based approach to generate MASes.
  We search for multi-agent systems in code space without relying on LLMs for code generation.
  The grammar is general-purpose and specifies constraints on how to combine components;
  considering additional base MASes only requires identifying their components.
  The search process generates candidate MASes;
  the best one is selected based on validation accuracy.
}

\end{figure*}
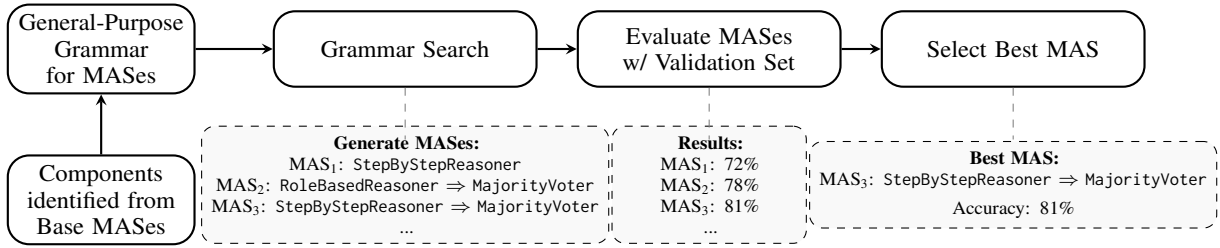
Despite their progress, existing automatic methods face key challenges.
Many methods such as ADAS \cite{hu2025automated} and AFlow \cite{zhang2025aflow}
operate directly in code space, which, although expressive,
often leads to syntactically invalid or semantically inconsistent programs
that waste computational resources.
In our analysis, ADAS generated invalid candidate MASes
in more than 20\% of search attempts.
Furthermore, automatically discovered MASes can become unnecessarily complex,
with large numbers of interconnected components that are difficult
to interpret or extend.
These issues limit the efficiency, transparency, and usability of MAS search.

To address these challenges, we introduce \textbf{Grammar Search},
a general framework for designing and discovering MASes that emphasizes
modularity, simplicity, and correctness.
The central idea is to represent MASes using a context-free grammar
that defines how individual components can be assembled
into larger systems.
These components can be manually designed, obtained by decomposing
existing MASes into smaller parts, or suggested by an LLM.
The resulting components are constrained by our grammar,
which encodes rules specifying how they can be connected.
Sampling MASes from this grammar guarantees syntactic correctness
and enables a structured search for novel MASes.
The grammar is fully extensible and not restricted to the initial set of components.
New components can be added as needed, making the framework adaptable
to new tasks and domains.

By defining a structured space of valid MASes,
Grammar Search effectively eliminates the need for LLM-based code generation
during search, thereby preventing invalid programs
and simplifying subsequent experimentation.
Although the components are simple,
the discovered MASes achieve competitive or superior performance
compared to previous automatic methods across multiple reasoning benchmarks.
At the same time, the search requires less computational effort,
and the resulting MASes remain more interpretable.

In summary, our contributions are as follows:\footnote{
    Code at: \url{https://github.com/mayanks43/grammar_search}.
}
\begin{compactenum}
    \item We propose \textbf{Grammar Search},
    a general framework for MAS discovery based on a context-free grammar
    that ensures modularity, correctness, and extensibility.
    \item We demonstrate that our framework outperforms
    existing automatic search methods on four out of five benchmarks
    across two domains: mathematics and question answering.
    \item Beyond accuracy, our framework provides a cheaper search stage
    and interpretable, simple, yet effective MASes.
\end{compactenum}

\section{Preliminaries}
\label{s:definitions}
We begin by defining a few key terms used throughout this work.

\paragraph{Agents}
Following \citet{plaat2025agenticlargelanguagemodels},
we define \textit{agents} as large language models (LLMs) that have been
adapted for specific tasks through prompting or post-training, and that can
\textit{reason} through intermediate steps toward goals, \textit{act} by executing
operations or invoking external tools, and \textit{interact} with users,
environments, or other agents through communication channels such as web
search, APIs, or memory modules.

\paragraph{Multi-Agent Systems}
A \textit{Multi-Agent System} (MAS) is a collection of agents coordinating toward
a shared goal.
MASes are commonly represented as graphs \cite{zhuge2024gptswarm}, where nodes
represent individual agents and edges represent channels of communication.
Alternatively, MASes can be represented through executable code
\cite{hu2025automated}, where the structure of function calls and data flow
explicitly defines the MAS.

\paragraph{MAS Search Space}
MASes span a large combinatorial search space that can vary along multiple axes,
including the types of nodes, their interconnections, and the objectives used to
evaluate them. 
We outline these axes at a conceptual level and provide examples from prior work
in Section~\ref{s:related_work}.

\section{Related Work}
\label{s:related_work}

\paragraph{Manually-designed MASes}
Most initial MASes were manually designed. Here, we describe some popular ones.  
Chain of Thought (CoT)~\cite{wei2023chainofthoughtpromptingelicitsreasoning}  
is often considered a trivial MAS, consisting of a single node with a prompt  
to solve the task step by step.  
Chain of Thought with Self-Consistency (CoT-SC)~\cite{wang2023selfconsistency}  
builds on CoT by using multiple parallel CoT nodes, followed by a  
majority-voting node that selects the most common answer  
among parallel outputs.  
Self-Refine~\cite{madaan2023self}  
uses the same model to iteratively refine an initial answer  
based on self-generated feedback.  
In Multi-Agent Debate~\cite{du2023improving},  
multiple agents debate over each other's answers,  
followed by a final decision agent that aggregates outputs  
from these agents.  
Tree-of-Thoughts~\cite{yao2023tree}  
solves problems by exploring a tree of possible next steps,
evaluating each branch, and backtracking from mistakes to find the optimal solution.
Instead of manually designing MASes, in this work we focus on  
automatically searching for them given the task.  
We use the first four MASes as baselines for comparison,  
as well as a basis for our MAS grammar.

\paragraph{Automatic MAS Search}
Lately, there has been a plethora of methods for automatically searching
for the best MASes.
These methods differ in how they represent, evaluate, and search for the best MASes:

\subparagraph{MAS Representation}
Different works have adopted different ways to represent MASes.  
Graph-based approaches such as GPTSwarm~\cite{zhuge2024gptswarm}
and MaAS~\cite{zhang2025multi} represent agents as nodes
and communication links as edges, allowing structural optimization of the graph.  
In contrast, code-based approaches such as ADAS~\cite{hu2025automated}
and AFlow~\cite{zhang2025aflow} represent MASes as executable Python programs,
where nodes correspond to API calls that can be connected in complex ways.
This representation allows fine-grained logic such as branching,
iteration, and function composition.
Our method follows this latter direction, using a grammar to define
composable code components that can be systematically explored.

\subparagraph{Node Design and Optimization}
Each node in a MAS can be optimized along multiple dimensions.  
Prompts can be refined or evolved through techniques such as PromptBreeder~
\cite{fernando2023promptbreederselfreferentialselfimprovementprompt},
or modified to encode distinct roles or personas for different agents
\cite{du2023improving}.
LLM parameters can also be adapted through post-training,
for example via fine-tuning as in Sirius~\cite{zhao2025sirius}.  
In addition, new tools can be dynamically created when existing ones are
insufficient for a task \cite{qiu2025alitageneralistagentenabling}.

\subparagraph{Search Frequency and Scope}
MAS search can occur at different levels of granularity.
Corpus-level search methods such as ADAS, AFlow, and GPTSwarm discover
one MAS that performs well across a dataset, while query-level methods
such as MaAS, FlowReasoner~\cite{gao2025flowreasoner},
and MasRouter~\cite{yue2025masrouter}
generate or adapt a new MAS for each input query.  
Corpus-level methods are generally more efficient and stable,
whereas query-level methods allow greater specialization.
To maintain efficiency, we focus on corpus-level search in this work.

\subparagraph{Search Objectives}
Search methods differ in their optimization objectives.
Common objectives include benchmark accuracy, API call cost, latency,
and safety considerations.
Many recent approaches such as AFlow
and MAS-ZERO~\cite{ke2025maszerodesigningmultiagentsystems}
adopt multi-objective optimization to balance these factors.
In this work, we focus on benchmark accuracy and API call cost
to discover MASes that achieve high accuracy while maintaining low search cost.

\subparagraph{Evaluation and Candidate Selection}
Candidate MAS evaluation strategies also vary across works.  
GPTSwarm and ADAS select candidates based on validation set performance,  
while MAS-ZERO uses prompting,
i.e., asking an LLM to choose the better MAS based on observed behavior.  
In this work, we evaluate candidate MASes using validation performance,
as it provides a more consistent and reliable measure of quality.

\subparagraph{Search Algorithms}
Several strategies have been explored to traverse the MAS space.
Iterative search (ADAS, MAS-ZERO) repeatedly asks an LLM to propose, refine,
and build upon existing candidates.
Reinforcement learning approaches such as GPTSwarm perform search and learning
in parallel.
Tree search variants such as Monte Carlo Tree Search (AFlow)
have also been explored for structured exploration of MAS graphs.
We adopt a random search strategy that explores the MAS space
component by component.

\begin{figure}[t]
\centering
\scriptsize
\renewcommand{\arraystretch}{1.2}
  \begin{tabular}{rcl}
$\langle System \rangle$ 
  & $\to$ & $\langle StartSI \rangle$ \\[0.6em]

$\langle StartSI \rangle$ 
  & $\to$ & $\langle StartSISO \rangle$ \\
  & $|$  & $\langle StartSISO \rangle\ \langle SI \rangle$ \\
  & $|$  & $\langle SIMO \rangle\ \langle MI \rangle$ \\[0.6em]

$\langle SI \rangle$ 
  & $\to$ & $\langle SISO \rangle$ \\
  & $|$  & $\langle SISO \rangle\ \langle SI \rangle$ \\
  & $|$  & $\langle SIMO \rangle\ \langle MI \rangle$ \\[0.6em]

$\langle MI \rangle$ 
  & $\to$ & $\langle MISO \rangle$ \\
  & $|$  & $\langle MISO \rangle\ \langle SI \rangle$ \\
  & $|$  & $\langle MIMO \rangle\ \langle MI \rangle$ \\[0.6em]

$\langle StartSISO \rangle$ 
  & $\to$ & \texttt{StepByStepReasoner[cnt=1]} \\
  & $|$  & \texttt{RoleBasedReasoner[cnt=1]} \\[0.6em]

$\langle SISO \rangle$ 
  & $\to$ & \texttt{StepByStepReasoner[cnt=1]} \\
  & $|$  & \texttt{RoleBasedReasoner[cnt=1]} \\
  & $|$  & \texttt{SelfCriticIteration[rnds=5]} \\[0.6em]

$\langle SIMO \rangle$ 
  & $\to$ & \texttt{StepByStepReasoner[cnt=5]} \\
  & $|$  & \texttt{RoleBasedReasoner[cnt=5]} \\[0.6em]

$\langle MISO \rangle$ 
  & $\to$ & \texttt{MajorityVoter} \\
  & $|$  & \texttt{ConsensusBuilder} \\[0.6em]

$\langle MIMO \rangle$ 
  & $\to$ & \texttt{DebateIteration[rnds=2]} \\
  & $|$  & \texttt{MultiSelfCriticIteration[rnds=5]} \\
\end{tabular}

\caption{Our general-purpose context-free grammar to represent MASes. Note: SI = SingleInput,
SO = SingleOutput, MI = MultiInput, MO = MultiOutput, cnt = count, and rnds = rounds.}
\label{fig:grammar}
\end{figure}

\paragraph{Summary and Relation to This Work}
Across methods described in this section, the core challenge is the same:
the MAS search space is vast.
Our work differs in that it introduces a context-free grammar to
define and constrain this space.
This provides a principled way to systematically enumerate, compare,
and analyze MASes while retaining the flexibility
of code-based representations.

\section{A Grammar for MASes}
\label{s:grammar}

In this section, we describe our framework in detail.  
A context-free grammar serves as the central element of our approach.  
This grammar defines the structured design space that the framework explores
when searching for valid MASes.  
We define a general-purpose grammar, shown in Figure~\ref{fig:grammar},
to enable systematic composition of MAS components.
The MAS components appear in the figure as the terminals of the grammar.  
To obtain these components, we manually decompose several existing MASes
into smaller subcomponents.
We will refer to these MASes as \emph{base} MASes to contrast them
with the automatically discovered MASes obtained through search.
By \emph{manually decompose}, we mean that we examine the code implementations
of these MASes and identify sections that can be encapsulated as
stand-alone components.
More specifically, we decompose four manually designed base MASes
into smaller components: CoT, CoT-SC, Self-Refine, and Multi-Agent Debate.
However, this decomposition is not limited to these four MASes
and can be extended to any group of MASes that differ substantially.  

\begin{table*}[t]
\centering
\small
  \begin{tabular}{l l l}
\toprule
\bf MAS & \bf Component Sequence & \bf Grammar Derivation \\
\midrule
CoT & 
\texttt{StepByStep[1]} &
$\begin{array}{@{}l@{}}
\langle \text{System} \rangle \to \langle \text{StartSI} \rangle \\
\langle \text{StartSI} \rangle \to \langle \text{StartSISO} \rangle \\
\langle \text{StartSISO} \rangle \to \texttt{StepByStep[1]}
\end{array}$ \\
\midrule
CoT-SC & 
\texttt{StepByStep[5]} $\Rightarrow$ \texttt{MajorityVoter} &
$\begin{array}{@{}l@{}}
\langle \text{System} \rangle \to \langle \text{StartSI} \rangle \\
\langle \text{StartSI} \rangle \to \langle \text{SIMO} \rangle\ \langle \text{MI} \rangle \\
\langle \text{SIMO} \rangle \to \texttt{StepByStep[5]} \\
\langle \text{MI} \rangle \to \langle \text{MISO} \rangle \\
\langle \text{MISO} \rangle \to \texttt{MajorityVoter}
\end{array}$ \\
\midrule
Self-Refine & 
\texttt{StepByStep[1]} $\Rightarrow$ \texttt{SelfCritic[5]} &
$\begin{array}{@{}l@{}}
\langle \text{System} \rangle \to \langle \text{StartSI} \rangle \\
\langle \text{StartSI} \rangle \to \langle \text{StartSISO} \rangle\ \langle \text{SI} \rangle \\
\langle \text{StartSISO} \rangle \to \texttt{StepByStep[1]} \\
\langle \text{SI} \rangle \to \langle \text{SISO} \rangle \\
\langle \text{SISO} \rangle \to \texttt{SelfCritic[5]}
\end{array}$ \\
\midrule
MA-Debate & 
\texttt{RoleBased[5]} $\Rightarrow$ \texttt{Debate[2]} $\Rightarrow$ \texttt{Consensus} &
$\begin{array}{@{}l@{}}
\langle \text{System} \rangle \to \langle \text{StartSI} \rangle \\
\langle \text{StartSI} \rangle \to \langle \text{SIMO} \rangle\ \langle \text{MI} \rangle \\
\langle \text{SIMO} \rangle \to \texttt{RoleBased[5]} \\
\langle \text{MI} \rangle \to \langle \text{MIMO} \rangle\ \langle \text{MI} \rangle \\
\langle \text{MIMO} \rangle \to \texttt{Debate[2]} \\
\langle \text{MI} \rangle \to \langle \text{MISO} \rangle \\
\langle \text{MISO} \rangle \to \texttt{Consensus}
\end{array}$ \\
\bottomrule
\end{tabular}

\caption{
  The four base MASes we work with,
  their component sequences,
  and their derivations with our grammar.
  We condense the names of some components to save space.
  \textit{Component sequence} depicts how the respective MAS is represented in our grammar.
  \textit{Derivation} shows how to derive the MAS through our grammar rules.
}
\label{t:reconstruction}
\end{table*}

The primary goal of our grammar is to enable the exploration of new MASes
composed of the smaller components derived from this decomposition.  
It is designed to be general, allowing new components to be added as needed.  
We now describe the grammar in detail.

\paragraph{Grammar Non-terminals}
The core of our grammar consists of \emph{non-terminals},
defined solely by their input-output structure.
We distinguish four categories: \textbf{SISO} (Single Input, Single Output),
\textbf{SIMO} (Single Input, Multiple Outputs),
\textbf{MISO} (Multiple Inputs, Single Output), and
\textbf{MIMO} (Multiple Inputs, Multiple Outputs),
each specifying the number of inputs and outputs they allow.

As might be evident, our grammar generally enforces that input
and output counts match between components when chained together.
Some components, such as \texttt{SelfCriticIteration} (which refines a provided answer),
cannot serve as the first component and therefore require additional logic.

Another constraint that enables seamless composition is
that each component must return either a single complete answer
or multiple complete answers. 
Subsequent components either build directly on the previous answers
or generate their own, using the prior answers as a side input.
This simple formulation allows any component of any MAS to naturally integrate
into our grammar.

\paragraph{Grammar Terminals}
The terminals in our grammar correspond to concrete components
that can be combined to construct new MASes.  
We now describe the major components obtained from decomposing
the four basic MASes mentioned earlier.  
However, our grammar is designed to be extensible and can easily
accommodate additional components beyond these.

\subparagraph{StepByStepReasoner(\texttt{count}=\textit{x})}

This is the Chain-of-Thought component. Given a task, the LLM is asked
to answer step by step. 
The \texttt{count} parameter allows the generation of multiple answers in parallel. 
This component may behave as either SISO or SIMO depending
on the number of answers generated.

\subparagraph{RoleBasedReasoner(\texttt{count}=\textit{x})}

This component functions similarly to StepByStepReasoner,
but with the added instruction that the LLM should assume a specific role or persona. 
For example, it may be asked to take on the role of a math professor
when solving math problems. 
The \texttt{count} parameter again enables the generation of multiple parallel answers. 
Depending on the value of \texttt{count}, this component may be either SISO or SIMO.

\subparagraph{SelfCriticIteration(\texttt{rounds}=\textit{x})}

This component consists of a critic agent followed by a reflect agent,
connected in sequence and repeated over a number of rounds as specified
by the \texttt{rounds} parameter. 
The critic agent critiques a response it receives from a previous component,
and the reflect agent uses the critic's feedback to generate a better answer. 
This is strictly a SISO component.

\subparagraph{DebateIteration(\texttt{rounds}=\textit{x})}

This is a MIMO component that requires input from a SIMO or another MIMO component. 
It performs iterative debates among multiple agents using the input answers it receives. 
The \texttt{rounds} parameter specifies how many iterations the debate process undergoes.

\subparagraph{MultiSelfCriticIteration(\texttt{rounds}=\textit{x})}

This is a MIMO version of SelfCriticIteration. 
It applies the critic-and-reflect sequence in parallel
across multiple input answer streams.

\subparagraph{MajorityVoter}

This is a MISO component that condenses multiple answer streams into a single answer. 
As the name suggests, it selects the most common answer among those provided.

\subparagraph{ConsensusBuilder}

This is another MISO component. 
Unlike MajorityVoter, it does not select an existing answer. 
Instead, the component synthesizes a new final answer based on the multiple answers
it receives as input.

\medskip
\noindent
Note that MultiSelfCriticIteration is not a component of the base MASes.
It was added as a representative example to demonstrate
how easily new components can be integrated into the grammar.

\medskip
\noindent
Table~\ref{t:reconstruction} shows representative sequences
in our grammar corresponding to the four initial MASes, along with their derivations.

\paragraph{MASes from component sequences}
Each terminal in our grammar corresponds to a code fragment.
These fragments follow a simple input-output interface:
they take the task details and a list of answers from the previous component
as input and return a list of answers as output.
The only exception is the first component,
which takes the task details as input. 
For single-input components, the input list contains a single answer;
likewise, the output list contains one answer for single-output components.

To construct a MAS corresponding to a component sequence,
the associated code fragments are stitched together,
with inputs and outputs passed along the sequence.  
Each sequence therefore corresponds to a deterministic program
representing a MAS.
Example sequences and their corresponding MASes are provided in
Appendix~\ref{a:appendix_4}.

\section{Searching Over MAS Grammar}
\label{s:grammar_search}

We use the grammar defined above to sample new component sequences
and select the best-performing corresponding MAS.
By randomly sampling a sequence, we mean starting from the start symbol
of the grammar and repeatedly choosing production rules at random
until a terminal-only sequence is obtained.

Our general algorithm proceeds as follows:

\begin{compactenum}
    \item Randomly sample a component sequence.
    \item If the sequence has not been seen before,
    evaluate the corresponding MAS on the validation set.
\end{compactenum}

After a fixed number of iterations, we select the best MAS based
on the performance scores of the sampled MASes on the validation set.

However, we find that pure random sampling often fails to include components
that are buried deeper in the grammar.  
To ensure that all components are fairly represented,
we adopt a \textit{forced sampling} strategy.

In forced sampling, we repeatedly sample sequences
until one containing a specific target component is found,
and then evaluate the resulting MAS on the validation set.  
We depict the full algorithm in Appendix~\ref{a:appendix_2} but also
summarize it here:
\begin{compactenum}
    \item Maintain a count of how many times each component has been sampled.
    \item Cluster components by their sampling counts.
    \item Begin with the lowest-count cluster (initially~0)
    and force-sample sequences containing each component in the cluster,
    followed by evaluation of the resulting MAS.
    \item Exhaust all components in the lower-count clusters
    before moving on to higher ones.
\end{compactenum}

Over time, if a component falls behind in its sampling frequency,
it will move into the lowest-count cluster and be force-sampled accordingly.  
This mechanism ensures that all components in the grammar are sampled
and evaluated in a fair and balanced manner.

We also explored more sophisticated search methods,
such as Thompson Sampling \cite{thompson1933likelihood}
and Monte Carlo Tree Search \cite{kocsis2006bandit},
to improve the discovery of the best component sequence.
However, we did not observe any performance gains from their use
and therefore chose to retain our simpler forced sampling strategy.

\section{Evaluation}
\label{s:evaluation}
\begin{table*}[t]
  \small
  \centering
  \begin{tabular}{@{} lrrrrrr @{} }
    \toprule
    & \multicolumn{3}{c}{\bf Math} & \multicolumn{2}{c}{\bf Other Domains} & \multirow{2}{*}{\bf Avg.}\\
    \cmidrule(lr){2-4} \cmidrule(lr){5-6}
    & \bf MATH500 &\bf  MATH lvl 5 & \bf AIME & \bf GPQA-D & \bf MMLU-Pro  \\\midrule
    
    Manually-Designed MASes \\

    \quad Chain of Thought & 77.7 ± 0.8 & 61.7 ± 1.7 & 7.6 ± 1.2 & 43.2 ± 2.5 & 65.9 ± 1.4 & 51.2 \\

    \quad CoT-Self Consistency & 79.3 ± 0.7 & 64.2 ± 1.3 & 8.5 ± 1.6 & 43.7 ± 2.6 & 67.1 ± 0.8 & 52.6 \\

    \quad Self-Refine & 78.1 ± 1.1 & 62.8 ± 1.2 & 8.2 ± 1.7 & 43.3 ± 2.2 & 66.2 ± 1.1 & 51.7 \\

    \quad Multi-Agent Debate & 79.8 ± 0.9 & 65.8 ± 1.1 & 10.4 ± 1.7 & 44.6 ± 2.5 & 67.4 ± 1.0 & 53.6 \\

    \midrule

    Automatic MAS Search Methods \\

    \quad ADAS & 80.9 ± 0.9 & 66.6 ± 1.4 & 8.7 ± 1.3 & 46.2 ± 2.3 & \textbf{68.5 ± 1.2} & 54.2 \\

    \quad AFlow & 80.0 ± 1.0 & 65.7 ± 1.0 & 8.6 ± 1.5 & 42.8 ± 2.9 & 65.4 ± 0.8 & 52.5 \\

    \quad Grammar search (Ours) & \textbf{81.4 ± 0.5} & \textbf{67.2 ± 1.6} & \textbf{11.1 ± 1.3} & \textbf{46.6 ± 2.0} & 68.4 ± 0.7 & \textbf{55.0} \\

    \bottomrule
\end{tabular}

  \caption{
    Results (accuracy and standard deviation across all runs) obtained with manually-designed MASes (top block)
    and three automatic MAS search methods (bottom block).
    Grammar Search, our approach,
    outperforms previous MAS search methods on average and in all but one benchmark.
    Additionally, Grammar Search never generates an invalid MAS and it is more efficient (Section \ref{s:evaluation}).
}
  \label{t:results}
\end{table*}

In this section, we discuss the evaluation setup and results for our method on various
datasets.

\paragraph{Datasets}

We evaluate our method on four datasets: two from mathematics and two from question answering.
We describe them below:

\textbf{MATH} \cite{hendrycks2021measuringmathematicalproblemsolving}:  
This dataset contains 12{,}000 competition-style math problems.  
For the search phase, we use 160 random problems from the training set as our validation set.  
For testing, we evaluate on two subsets: the 500-problem partition known as MATH500,
and a separate partition of 486 level-5 problems, following the setup in AFlow.

\textbf{AIME} \cite{aime_1983_2024}:  
This dataset consists of questions from the American Invitational Mathematics Examination
and is more challenging than the MATH dataset.  
We use 160 random problems from the 1983--2019 set for validation.  
For testing, we use 133 problems from the years 2020--2024.  
We avoid the more common 30-problem split from the most recent year
to ensure greater statistical power during evaluation \cite{hochlehnert2025sober}.  

\textbf{MMLU-Pro} \cite{wang2024mmlu}:  
MMLU-Pro is a more difficult variant of the MMLU benchmark \cite{hendryckstest2021},
a general-domain QA dataset.  
Unlike the original MMLU, which provides 4 answer choices per question, MMLU-Pro
includes 10 options per question.  
Because the provided validation set contains only 70 problems,
we construct larger validation and test sets by splitting the full test set into two parts:
one for validation and one for testing.  
From the validation split, we randomly select 160 problems for the validation set, and from the test split, we randomly select 500 problems for the test set.
Some problems had to be dropped due to content policy triggers from our LLM API provider.
As a result, the final validation and test sets contain 158 and 494 problems, respectively.
To the best of our knowledge, we are the first to evaluate automatic MAS search methods on this dataset.

\textbf{GPQA} \cite{rein2023gpqagraduatelevelgoogleproofqa}:  
The Google-Proof Question Answering dataset (GPQA) contains expert-authored questions
across multiple disciplines, designed to resist shallow, web-search-based answering.  
As our test set, we use the hard variant GPQA-Diamond, which contains 198 questions.  
To construct the validation set, we subtract the 198 Diamond problems from the main GPQA set,
yielding 250 remaining problems.  
From these, we randomly sample 160 for the validation set.  

\paragraph{Baselines}
As baselines, we use the four base MASes we have been working with:
CoT, CoT-SC, Self-Refine, and Multi-Agent Debate.
In addition, we compare our method against two leading corpus-level
automatic MAS search methods: AFlow and ADAS.
We run ADAS for 30 iterations and AFlow for 20 iterations,
following the default settings in their respective codebases.
Similar to ADAS, we run our search algorithm for 30 iterations on all evaluated datasets.

\paragraph{Evaluation Metric}
The evaluation metric for all datasets is accuracy.
Following recent works that employ LLMs for answer equivalence checking
\citep{ho2025llm, ke2025mas},
we consider a prediction correct if it matches the ground truth exactly
or is judged equivalent by a strong LLM (\texttt{gpt-5}; \citealt{openai2025gpt5})
when a direct equality check fails.

For all datasets, following \citet{zhang2025aflow},
we evaluate each problem five times on the validation set
during the search phase and use the average accuracy to guide the search for the best MAS.
In the test phase, we evaluate the best MAS found
during the search phase eight times on the test sets of MATH and MMLU-Pro
and report the average accuracy and standard deviation.
Because the GPQA and AIME test sets are smaller,
we evaluate them 20 and 30 times, respectively,
following the recommendations of \citet{hochlehnert2025sober}
for repeated evaluations.
These numbers were chosen to keep the total number of evaluations
roughly consistent with those of MATH and MMLU-Pro.

\paragraph{LLM Setup}  
Throughout our evaluations, we use \texttt{gpt-4o-mini} as the backbone LLM
(i.e., the LLM powering the nodes in the MAS).  
If a method requires the use of an optimizer LLM or a meta-agent
(i.e., the LLM that proposes the next MAS to evaluate during the search phase),
we use \texttt{gpt-4.1-mini}, as it is a newer and stronger model than \texttt{gpt-4o-mini}.  
Note that our method does not require an optimizer LLM.

\paragraph{Results}
We now discuss the results obtained by evaluating our method under the setup described above.  
The main results are shown in Table~\ref{t:results}.  
We find that our method performs better than
the next-best automatic search method, ADAS,
on four out of five evaluated datasets as well as on average.
Our method improves accuracy by 0.8 points over ADAS, 2.5 points over AFlow,
and 1.4 points over the strongest manually designed MAS baseline (Multi-Agent Debate).
On AIME in particular, our method achieves a 2.5\% absolute improvement
over other automatic search methods.
In addition, we find that our method tends to produce lower standard deviations, with an average value of approximately 1.0 compared to 1.2 for ADAS, 1.2 for AFlow, and 1.4 for manually designed MASes.
This indicates that Grammar Search not only performs better on average
but also yields more stable and reliable MASes.  

\subsection{Evaluation Beyond Accuracy}
Accuracy is not the only factor that matters when choosing a framework.
Therefore, we further compare Grammar Search and ADAS on metrics
related to efficiency and MAS quality, as summarized in
Table~\ref{t:cost_results}.
Grammar Search is more efficient, reliable, and interpretable
than ADAS while achieving comparable accuracy.
Our framework eliminates wasted computation during MAS generation,
reduces overall API costs by approximately 12\%,
and always produces simpler, valid MASes.
While ADAS occasionally discovers a strong MAS early in the search,
it spends substantial computation on invalid or overly complex MASes,
whereas Grammar Search reliably constructs valid and concise MASes
through structured component composition.
Note that by \emph{valid} MASes,
we mean those that do not contain any syntactic or runtime errors
and achieve an accuracy greater than zero when evaluated on the validation set.  

During the search phase, Grammar Search incurs zero cost for MAS generation
because it does not rely on LLMs to generate code.
All MASes are constructed from verified components, ensuring 100\% validity
and eliminating wasted computation on invalid MASes.
By contrast, ADAS incurs nontrivial generation costs
(\$0.06 per valid MAS and \$0.25 per invalid MAS)
and wastes computation on invalid generations, which occur in 23\% of cases.
For the evaluation of valid MASes, Grammar Search and ADAS perform
a similar number of trials, leading to comparable evaluation costs.
Overall, Grammar Search achieves an average total search cost
of \$8.37 per MAS, compared to \$10.25 for ADAS, a reduction of about 12\%.  

Beyond cost, Table~\ref{t:cost_results} also highlights major differences
in MAS reliability and complexity.
ADAS produces MASes with longer code (8,612 vs.\ 4,392 characters on average)
and substantial code growth across search iterations
(524 vs.\ 33 characters per iteration),
yet this added complexity does not lead to higher accuracy.
Moreover, ADAS tends to find its best MAS early (around iteration~3),
suggesting that later iterations mainly generate redundant or inefficient code.
In contrast, Grammar Search maintains steady exploration
and discovers its best MAS later (around iteration 11.5),
indicating a more systematic and stable search process.  

Overall, Grammar Search achieves lower search-phase cost,
guarantees valid MAS generation, and produces shorter,
more interpretable MASes than ADAS, demonstrating that structured,
grammar-based search can match or surpass the effectiveness
of free-form code generation while being substantially more efficient and reliable.
\begin{table}[t]
  \small
  \centering
  \begin{tabular}{@{}lrr@{}}
\toprule
& \bf ADAS & \bf Grammar \\
\midrule

Valid MASes                & 77\% & \textbf{100\%} \\
\quad Avg. generation cost & \$0.06 & \textbf{\$0.00} \\
\quad Avg. validation cost & \$10.19 & \textbf{\$8.37} \\
\quad Avg. total cost      & \$10.25 & \textbf{\$8.37} \\ \addlinespace

Invalid MASes & 23\% & \textbf{0\%} \\
\quad Avg. generation cost & \$0.25 & \bf \$0.00 \\
\quad Avg. validation cost & \$0.96 & \bf \$0.00 \\
\quad Avg. total cost & \$1.21 & \bf \$0.00 \\ \midrule

All MASes \\
\quad Avg. code length. (\#chars.) & 8,611.8 & \textbf{4,391.5} \\
\quad Avg. code growth  (\#chars.) &   523.8 & \textbf{32.8} \\ \midrule

Best MAS found at iteration & 3.0 & 11.5 \\

\bottomrule
\end{tabular}
  \caption{
    Analysis of MASes generated during the search phase of ADAS and Grammar Search.    
    Code growth is measured with respect to the previously generated MAS.
    23\% of ADAS-generated MASes are invalid (the code does not run or they obtain 0\% accuracy).
    Grammar Search is more efficient across the board,
    and the generated MASes grow at a slower pace and are shorter.
  }
  \label{t:cost_results}
\end{table}
\begin{table}[t]
\centering
\small
\begin{tabular}{lcc}
\toprule
\textbf{Dataset} & \textbf{Forced} & \textbf{Random} \\
\midrule
MATH     & 81.6 ± 0.7  & 80.8 ± 0.7 \\
AIME     & 24.6 ± 2.2  & 24.0 ± 1.6 \\
GPQA     & 41.7 ± 3.0  & 41.6 ± 2.6 \\
MMLU-Pro & 70.1 ± 1.8  & 69.5 ± 2.1 \\
\midrule
\textbf{Avg.} & 54.5 ± 2.0 & 54.0 ± 1.8 \\
\bottomrule
\end{tabular}

\caption{
  Validation accuracies of the best MASes found by Grammar Search with forced and random sampling.
  Random sampling performs worse on average.
}
\label{t:forced_vs_random}
\end{table}
\paragraph{Best MASes found}
We depict the best MASes found through our method in Appendix~\ref{a:appendix_1}.
Across datasets, the component sequences share several components,
most notably DebateIteration and StepByStepReasoner (the SIMO version),
suggesting that these are generally strong building blocks.

\paragraph{Ablation}
We also describe an ablation experiment comparing the forced sampling strategy
with the random sampling strategy for sampling component sequences during the search phase.  
Table~\ref{t:forced_vs_random} presents the results of this ablation.  
While the gains are not large, they are consistent across datasets,
indicating a clear advantage of the forced sampling strategy over random sampling.

\section{Conclusion}
\label{s:conclusion}

In this work, we proposed \textbf{Grammar Search}, a new corpus-level framework
for automatic MAS discovery in code space.
Our method constructs a context-free grammar that defines a structured
and extensible design space for composing multi-agent systems.
This grammar constrains how building-block components can be combined,
ensuring that only valid MASes are generated.
By sampling from the grammar, we produce valid MAS candidates during the search phase,
thereby avoiding wasted computation and API costs on invalid generations.
To maintain coverage and prevent under-testing, we employed a simple forced-sampling
strategy that balances the frequency of each component across sampled sequences.

We showed that Grammar Search outperforms existing automatic MAS search methods
and manually designed baselines on four out of five benchmarks,
as well as on average.
Furthermore, the MASes produced by our framework are modular,
interpretable, and efficient, providing simpler yet effective
multi-agent systems while reducing search cost.

\section*{Limitations}
\label{s:limitations}

This work has several limitations.  
First, our approach lacks the free-form expressivity of LLM-based code generation,
and the MASes it produces are constrained by the grammar we developed.
However, this constraint offers practical benefits, as it enables a cheaper search stage
and avoids wasted computation on invalid MASes.  
In addition, our grammar is extensible and could potentially be modified,
for example by an LLM, to approach the expressivity of other methods.  

Second, the best MASes generated by our method depend on the initial
components defined in the grammar.
If these components are weak, the resulting MASes will not achieve
performance comparable to that of the best existing methods.

Finally, although we were able to achieve good results with a relatively simple
random search strategy, we expect that expanding the grammar with more components
will require more sophisticated search algorithms
such as Monte Carlo Tree Search or Thompson sampling.

\section*{Ethics}
\label{s:ethics}

We work with LLMs in this study, which are known to carry various social and
ethical risks \cite{weidinger2021ethicalsocialrisksharm}.
Techniques such as RLHF have helped mitigate some of these risks,
but unsafe behaviors may still arise due to methods like LLM jailbreaking.

All datasets, models, and code used in this study were obtained
from public sources and used in accordance with their respective licenses
and terms of use.

\section*{Note on AI Assistant Usage}
\label{s:ai_assistant}

We acknowledge the use of AI tools such as GPT-5 and Claude Opus 4.1
for grammar checking, paraphrasing, and polishing the manuscript
to improve clarity, as well as for assistance
in writing and debugging portions of the associated code.

\bibliography{custom}

\appendix

\section{Appendix 1}
\label{a:appendix_1}
The best component sequences discovered across each dataset using Grammar Search
are depicted in Table~\ref{t:best_mas_sequences}.

\begin{table*}[t]
  \centering
  \resizebox{\linewidth}{!}{%
\begin{tabular}{@{}ll@{}}
\toprule
\textbf{Dataset} & \textbf{Best component sequence found} \\
\midrule
AIME & StepByStepReasoner(5) $\rightarrow$ DebateIteration(2) $\rightarrow$ DebateIteration(2) $\rightarrow$ MajorityVoter \\
GPQA & StepByStepReasoner(5) $\rightarrow$ MultiSelfCriticIteration(5) $\rightarrow$ DebateIteration(2) $\rightarrow$ ConsensusBuilder \\
MATH & StepByStepReasoner(5) $\rightarrow$ DebateIteration(2) $\rightarrow$ ConsensusBuilder \\
MMLU-Pro & StepByStepReasoner(5) $\rightarrow$ DebateIteration(2) $\rightarrow$ MajorityVoter $\rightarrow$ RoleBasedReasoner(1) $\rightarrow$ StepByStepReasoner(1) \\
\bottomrule
\end{tabular}
}
  \caption{
    We present the best component sequence discovered for each dataset using Grammar Search with the forced sampling strategy.  
    For brevity, parameter names are omitted.  
  }
  \label{t:best_mas_sequences}
\end{table*}

\section{Appendix 2}
\label{a:appendix_2}

Simplified python code for Grammar Search with Forced Sampling can be found in
Listing~\ref{lst:grammar_search}.

\begin{figure*}
\begin{lstlisting}[language=Python, caption={Grammar MAS Search with Forced Sampling}, label=lst:grammar_search]
def grammar_mas_search(grammar, validation_set, iterations):
    """
    Grammar-based Multi-Agent System search with forced component sampling.
    
    Args:
        grammar: Grammar rules for MAS generation
        validation_set: Validation dataset for evaluation  
        iterations: Total number of iterations
    
    Returns:
        best_sequence: Best component sequence found
    """
    # Initialize component counts and evaluated sequences
    component_counts = {c: 0 for c in get_all_components(grammar)}
    evaluated_sequences = set()
    sequence_scores = {}
    
    for t in range(1, iterations + 1):
        # Find components with minimum observation count
        min_count = min(component_counts.values())
        components_with_min_count = [c for c, count in component_counts.items() 
                                     if count == min_count]
        
        # Force each minimum-count component
        for component in components_with_min_count:
            # Sample sequences until target component appears
            while True:
                # expand_grammar(): randomly derive terminal sequence from grammar
                sequence = expand_grammar(grammar)  
                
                # Check if component in sequence and sequence is new
                if component in sequence and tuple(sequence) not in evaluated_sequences:
                    break
            
            # evaluate_sequence(): run sequence on validation set, return accuracy
            score = evaluate_sequence(sequence, validation_set)  
            
            # Update tracking structures
            evaluated_sequences.add(tuple(sequence))
            sequence_scores[tuple(sequence)] = score
            component_counts[component] += 1
    
    # Return sequence with highest score
    best_sequence = max(sequence_scores.keys(), key=lambda s: sequence_scores[s])
    return list(best_sequence)

\end{lstlisting}

\end{figure*}

\section{Appendix 3}
\label{a:appendix_3}
Our code infrastructure for calling LLMs inside MASes is similar to ADAS. 
A simplified version is depicted in
Listing~\ref{lst:core_components}.

\begin{figure*}[ht]
\begin{lstlisting}[language=Python, caption={Core Infrastructure Components}, label=lst:core_components]
# Named tuple for information passing between agents
from collections import namedtuple
Info = namedtuple('Info', ['name', 'author', 'content', 'iteration_idx'])

class LLMAgentBase:
    """Base class for LLM agents."""
    def __init__(self, agent_name, role='assistant', temperature=0.5):
        self.agent_name = agent_name
        self.role = role
        self.temperature = temperature
    
    def __call__(self, inputs, instruction, iteration_idx=-1):
        # Construct prompt from inputs and instruction
        prompt = self.format_inputs(inputs) + instruction
        # Query LLM and return Info object
        response = query_llm(prompt, self.temperature)
        return Info('answer', self.agent_name, response, iteration_idx)

class AgentSystem:
    """Base class for multi-agent systems."""
    def forward(self, taskInfo):
        # To be implemented by generated code
        pass
\end{lstlisting}
\end{figure*}

\section{Appendix 4}
\label{a:appendix_4}

We depict code for several MASes generated through our framework in
Listings~\ref{lst:cot_sc},~\ref{lst:self_refine} and \ref{lst:debate}.

\begin{figure*}[ht]
\begin{lstlisting}[language=Python, caption={Generated MAS: Self-Consistency with Voting}, label=lst:cot_sc]
def forward(self, taskInfo):
    # Component function definitions
    def component_0_step_by_step_plural(taskInfo, prev_answer=None):
        '''Multiple parallel step-by-step reasoners'''
        inputs = [taskInfo]
        instruction = "Please think step by step and then solve the task. Put your final answer in \\boxed{}."
        
        if prev_answer is not None:
            inputs.append(prev_answer)
            instruction = "Based on the previous solution above, please think step by step and provide your own solution. Put your final answer in \\boxed{}."
        
        N = 5
        agents = [LLMAgentBase('Chain-of-Thought Agent', temperature=0.8) for _ in range(N)]
        
        all_results = []
        for i in range(N):
            answer = agents[i](inputs, instruction)
            all_results.append(answer)
        
        return all_results, agents
    
    def component_1_majority_voter(taskInfo, all_results):
        '''Apply semantic-aware majority voting using LLM'''
        voting_instruction = "Given these " + str(len(all_results)) + " solutions to the same problem:\\n\\n"
        for i, response in enumerate(all_results, 1):
            voting_instruction += "\\nSolution " + str(i) + ":\\n" + response.content + "\\n"
        
        voting_instruction += "\\nAnalyze these solutions and identify which answer appears most frequently.\\n"
        voting_instruction += "Copy that ENTIRE solution verbatim, including all reasoning steps.\\n"
        voting_instruction += "After copying the solution, ensure your final answer is in \\boxed{}."
        
        voting_agent = LLMAgentBase('Voting Agent', temperature=0.1)
        final_answer = voting_agent([taskInfo], voting_instruction)
        
        return final_answer
    
    # Orchestration
    all_results, agents = component_0_step_by_step_plural(taskInfo)
    answer = component_1_majority_voter(taskInfo, all_results)
    return answer
\end{lstlisting}
\end{figure*}

\begin{figure*}[ht]
\begin{lstlisting}[language=Python, caption={Generated MAS: Single Reasoner with Self-Criticism}, label=lst:self_refine]
def forward(self, taskInfo):
    # Component function definitions
    def component_0_step_by_step_singular(taskInfo, prev_answer=None):
        '''StepByStepReasoner: Chain-of-Thought reasoning with single agent'''
        inputs = [taskInfo]
        instruction = "Please think step by step and then solve the task. Put your final answer in \\boxed{}."
        
        if prev_answer is not None:
            inputs.append(prev_answer)
            instruction = "Based on the previous solution above, please think step by step and provide your own solution. Put your final answer in \\boxed{}."
        
        agent = LLMAgentBase('Chain-of-Thought Agent')
        answer = agent(inputs, instruction)
        return answer, agent
    
    def component_1_self_critic_plural(taskInfo, answer, agent=None):
        '''Pure refinement iteration - multiple rounds'''
        if agent is None:
            agent = LLMAgentBase('Refinement Agent')
        
        inputs = [taskInfo]
        
        critic_instruction = "Please review the answer above and provide detailed feedback on any errors or improvements needed. "
        critic_instruction += "At the end of your feedback, write either [CORRECT] or [INCORRECT]."
        
        critic_agent = LLMAgentBase('Critic Agent')
        
        reflect_instruction = "Given previous attempts and feedback, carefully consider where you could go wrong. "
        reflect_instruction += "Using insights from previous attempts, try to solve the task better. Put your final answer in \\boxed{}."
        
        N_max = 5
        for i in range(N_max):
            feedback = critic_agent([taskInfo, answer], critic_instruction, i)
            if '[CORRECT]' in feedback.content:
                break
            inputs.extend([answer, feedback])
            answer = agent(inputs, reflect_instruction, i + 1)
        
        return answer
    
    # Orchestration
    answer, agent = component_0_step_by_step_singular(taskInfo)
    answer = component_1_self_critic_plural(taskInfo, answer, agent)
    return answer
\end{lstlisting}
\end{figure*}

\begin{figure*}[ht]
\begin{lstlisting}[language=Python, caption={Generated MAS: Role-Based Debate with Consensus}, label=lst:debate]
def forward(self, taskInfo):
    # Component function definitions
    def component_0_role_based_plural(taskInfo, prev_answer=None):
        '''Multiple role-based agents with different perspectives'''
        inputs = [taskInfo]
        instruction = "Please think step by step and then solve the task. Put your final answer in \\boxed{}."
        
        if prev_answer is not None:
            inputs.append(prev_answer)
            instruction = "Based on the previous solution above, please think step by step from your role perspective and provide your own solution. Put your final answer in \\boxed{}."
        
        roles = ['Math Professor', 'Grade School Teacher', 'Math Enthusiast', 'Research Scientist', 'Teaching Assistant']
        agents = [LLMAgentBase('Role-Based Agent', temperature=0.8, role=role) for role in roles]
        
        all_results = []
        for agent in agents:
            answer = agent(inputs, instruction)
            all_results.append(answer)
        
        return all_results, agents
    
    def component_1_debate_plural(taskInfo, all_results, agents):
        '''Pure iteration loop where agents debate over multiple rounds'''
        iteration_instruction = "Given solutions to the problem from all agents (including yourself), "
        iteration_instruction += "consider all perspectives and provide an updated solution and answer. "
        iteration_instruction += "Put your final answer in \\boxed{}."
        
        N_max = 2
        for round_num in range(N_max):
            current_results = []
            for i in range(len(agents)):
                answer = agents[i]([taskInfo] + all_results, iteration_instruction)
                current_results.append(answer)
            
            all_results = current_results
        
        return all_results, agents
    
    def component_2_consensus_builder(taskInfo, all_results):
        '''Final decision-making by synthesizing all solutions'''
        final_instruction = "Given all the above solutions, analyze them carefully and provide a final solution and answer. "
        final_instruction += "Put your final answer in \\boxed{}."
        final_agent = LLMAgentBase('Final Decision Agent', temperature=0.1)
        
        answer = final_agent([taskInfo] + all_results, final_instruction)
        return answer
    
    # Orchestration
    all_results, agents = component_0_role_based_plural(taskInfo)
    all_results, agents = component_1_debate_plural(taskInfo, all_results, agents)
    answer = component_2_consensus_builder(taskInfo, all_results)
    return answer
\end{lstlisting}
\end{figure*}

\end{document}